\documentclass[11pt,letterpaper]{article}
\usepackage[hyphens]{url}
\usepackage{cogsys}

\usepackage{hyperref}  
\usepackage[T1]{fontenc}
\usepackage{xcolor}
\usepackage{times}
\usepackage{framed}
\usepackage{booktabs}
\usepackage{multirow}
\usepackage[final]{microtype}
\usepackage[pdftex]{graphicx} 

\usepackage[style=apa,backend=biber]{biblatex}
\newcommand{\sr}{\rule[-0.25cm]{0pt}{0.7cm}}

\cogsysheading{X}{20XX}{1-6}{X/20XX}{X/20XX}

\ShortHeadings{Computational Intentionality}
              {W.\ Bridewell}

\date{\today}

\begin{document}

\title{Taking the Intentional Stance Seriously, or \\ ``Intending'' to Improve Cognitive Systems} 
 
\author{Will Bridewell}{will.bridewell@nrl.navy.mil}
\address{U.S. Naval Research Laboratory,
         Washington, DC 20375 USA}
\vskip 0.2in
\begin{abstract}
    Finding claims that researchers have made considerable progress in artificial intelligence over the last several decades is easy. However, our everyday interactions with cognitive systems (e.g., Siri, Alexa, DALL-E) quickly move from intriguing to frustrating. One cause of those frustrations rests in a mismatch between the expectations we have due to our inherent, folk-psychological theories and the real limitations we experience with existing computer programs. The software does not understand that people have goals, beliefs about how to achieve those goals, and intentions to act accordingly. One way to align cognitive systems with our expectations is to imbue them with mental states that mirror those we use to predict and explain human behavior. This paper discusses these concerns and illustrates the challenge of following this route by analyzing the mental state `intention.' That analysis is joined with high-level methodological suggestions that support progress in this endeavor. 
\end{abstract}

\section{Introduction}
Largely due to commercial interest and support there has been a dramatic shift from projects in artificial intelligence (AI) that are mind inspired and theory driven toward ones that are brain inspired and data driven. In the span of a decade, machine learning approaches that draw inspiration from neuroscience and mathematics eclipsed research informed by psychology, philosophy, and economic theory. Even before deep learning dominated the field, researchers in AI replaced structured representations of intentional states such as belief, desire, intention, and choice with mathematical representations that provide formal guarantees on inference and learning. In the process, beliefs became probability distributions, actions became the outputs of learned policies, and rationality was equated with an optimality criterion. Simultaneously, the internals of models and systems in AI became increasingly opaque to the point that predicting and explaining their behavior became impractical, if not impracticable. 

In response, there has been a call for explainable AI (XAI) from the Defense Advanced Research Projects Agency (DARPA) to ``enable human users to understand, appropriately trust, and effectively manage the emerging generation of artificially intelligent partners'' \parencite{darpaXAI}.  In practice, researchers take this statement to mean that a causal explanation of an AI system's output is necessary for people to assign credence and to diagnose failures \parencite{Gunning2019,druce21}.  However, the imagined human-machine partnerships inspiring this call extend beyond deciding whether to accept machine responses and determining how to improve machine accuracy. Advice on this point is given by Clancey and Hoffman \parencite*{Clancey2021} and Hoffman and colleagues \parencite*{ihmc2021}. 

Looking at how we work with other people can provide a sense of what might be required for successful collaborations with AI partners. When we jointly act with people, we rely heavily on the application of folk-psychological theories: our understanding of intentional states like knowledge, beliefs, desires, intentions, hopes, and concerns. Taking what Dennett \parencite*{dennett87} calls the \textit{intentional stance}, involves using these theories to guide prediction, explanation, and planning so that we can navigate our social relationships.\footnote{The word `intentional' is used in two ways in this paper. When referring to the intentional stance, the reference is to intentionality, which is philosophical jargon for mental states being about some form of content. Beliefs, desires, hopes, wishes, and so on are about some other subject (e.g., I believe that X, I hope that X). In contrast, intentional action most often refers to planning to do something (e.g., I intend to go to the store later today, I intend to learn calculus) or the act of doing something on purpose (e.g., You intentionally spilled my drink!). Although there is a possibility of confusion, the intended meaning should be clear in context.} Contrast that with how we view our interactions with ordinary machines like toasters or televisions. In those cases, Dennett points out that we usually take a \textit{design stance}, which is informed by our knowledge of the functions those machines were built to carry out. We describe a toaster by saying that you put your bread in the slots, press down on the lever, and after some time your toast is ready. This description does not refer to mental states any more than it refers to physical laws. All that matters for prediction, explanation, and planning is a knowledge of the toaster's function. However, as intelligent machines become general purpose surrogates for people, the potential behaviors and interactions will increase in complexity to the point that the design stance becomes unmanageable for everyday interactions. The functionality of the machines and the rules guiding their behavior will become too complex to anticipate or learn. We see this already with intelligent systems like Amazon's Alexa and Apple's Siri, which are meant to simulate limited forms of human interaction and are constantly growing more capable. Furthermore, people not only naturally apply their folk-psychological concepts when explaining robotic behavior \parencite{DeGraaf2019a} but will also forfeit their careers to hold on to that interpretation \parencite{tiku22a,tiku22b}. 


I assume that easing human-machine collaboration requires closing the gap between the behavior of our artificially intelligent partners and the expectations we form from our folk-psychological theories. Likewise, de Graaf and Malle \parencite*{DeGraaf2017a} argue that explanations given by an AI system need to be couched in the terms of folk psychology for people to accurately calibrate their trust in that system \parencite{theodorou16}.  If this is the case, then the AI community needs to determine or specify the extent to which its systems align with those theories. After a brief digression on how the intentional stance is viewed within AI, I examine intention and its relationship to intentional action in detail to illustrate the challenges of accepting this assumption. Finally, I discuss how to evaluate AI systems that include artificial intention and how scientific progress toward a full understanding of intention can be made. 

\section{Searching for Intentionality in Modern AI}
Although AI has been through a radical shift in focus, the move to deep learning and neural network approaches has not eliminated taking the intentional stance. We still describe AlphaGo as \textit{wanting} to win a game and \textit{intentionally} placing stones in locations that it \textit{believes} are most strategic. That there are no structural analogs to the intentions, beliefs, or desires within the system is irrelevant. When Lee Se-dol was playing AlphaGo, his move predictions could not be based on the actual implementation and could only be driven by his assessment of what a rational opponent, trying to win, would do. And, when explaining a move by AlphaGo that surprised Lee (because it appeared irrational to him), there is a natural inclination to use a variety of intentionality-laden words to describe the process, including `discover' and `understand' \parencite{metz16}.  Such language tends to be used loosely and metaphorically, which serves to mask the inadequacy of the explanation. Readers are left to infer how the words apply, and whether they are experts or not, misunderstandings are unavoidable.

Some AI researchers might react to the loose use of the language of intentional states by resorting to formalization. This is the strategy behind the development of Belief-Desire-Intention (BDI) frameworks, which provide definitions of belief, desire, and intention and the sort of guarantees on rationality that accompany normative theories \parencite{Rao1995}. These frameworks are used to guide system design and to let people both predict behavior and explain how an agent's desires drive its behavior, at least under the aegis of a normative rationality. Building a system that adheres to such a theory of how intelligent systems should, in principle, behave does not imply that such systems characterize or describe how humans do behave. Moreover, capturing some of the characteristics of beliefs, desires, and intentions in a logic does not imply capturing all relevant characteristics that people bring to bear in their folk-psychological theories. For instance, knowing someone's plans facilitates the prediction of what kinds of objects, actions, and events will be important to them in pursuit of that plan, but there may be no such relationship in a system built using a BDI framework. When describing BDI agents, we may use the language of intentionality, but the terms have taken a meaning related to but not nearly as rich as their use in folk psychology. 

How then should we use the language of intentionality to communicate the way in which an intelligent system operates? The answer is rooted in empirical methodology. The next section delineates some of the functions and properties of intentional action. This list serves two roles. First, showing that a single intentional state is multifaceted acts as an argument that the referents of terms like `intention' and `belief' are more complex than is traditionally assumed in computational research. Second, the list provides, at a high level, target phenomena or capabilities that can be used when testing and evaluating a system. Importantly, the list is not a comprehensive accounting of functions and properties. As with attempts to achieve behavior as complex as autonomous driving, establishing a guide for system design and testing requires a serious and sustained interdisciplinary effort \parencite{ndot2018,webb2020}. The result would be the selection of desirable features, the development of corresponding test scenarios, and agreed upon levels of functionality. The last of these would enable people to align their expectations with the capabilities of an intelligent system.

\section{A Closer Look at Intentional Action}
\label{sec:lookia}
When researchers talk about intentional states or other psychological concepts, they typically carve up their characteristics so that they are only talking about them from a limited perspective. Sometimes researchers take these limited perspectives to be the core character of the concept within a specific, circumscribed scope. Consider belief. We might say that someone would be surprised if they (a) believed that a toaster was plugged in, (b) followed the normal directions for toasting bread, and (c) observed that their bread had not been toasted. The expectation based on the toaster's design has been violated. This relationship between belief and surprise expresses an important characteristic of belief but says nothing about other aspects such as strength of conviction, effects on mood, changes in expectations, or how the experience warrants other beliefs. Therefore, it would be premature to declare that a system that connects these thin representations of beliefs, expectations, and surprise to model this case has beliefs in a folk-psychological sense. To examine the complexity of intentional states and to motivate a testing and evaluation methodology, let us take a closer look at intention. 

Within AI, intention and resultant intentional action were initially analyzed in the context of speech acts \parencite{Allen1980a} and were eventually associated with planning \parencite{Bratman1988a}.  This emphasis was due largely to the influence of Bratman \parencite*{Bratman1984} who worked to shift attention from what he called present-directed intention, which supports intentional action in the moment, toward future-directed intention, which is a commitment to a later action. To illustrate the difference, the law is particularly concerned with intentional action and distinguishes among crimes planned in advance (i.e., premeditated), crimes of passion whose outcomes were intended only in the moment, and crimes of negligence where the negative effects were unintended. Although the relationship between plans and intentions had been discussed since the early days of cognitive science \parencite{miller60},  the connection remained unexploited in AI until the 1980s and eventually led to the development of BDI frameworks and agent-oriented programming \parencite{Shoham1993}.  

The relationship to planning is only one aspect of intention, and I will examine intention from three different perspectives to appreciate its richness. In addition to intention's connection with plans, I will consider its connection to both action and control. 

\subsection{Intention Through the Lens of Planning}

I have already mentioned the high-level intersection between intentions and plans, and here I highlight three characteristics that Bratman discusses in detail. The first is that intentions enable planning by storing the results of deliberation about which actions to take in the future. Without this ability, there would be no connection between plan creation at time $t_0$ and actions at later time $t_n$. This is not to say that intentions always exist as a fully specified set of actions. Steps in plans can be more or less abstract in the sense of, ``I intend to go to the bakery later,'' or they may include concrete times, locations, and subplans. Intentions can also be incomplete or open ended such as, ``I intend to buy a loaf of rye bread and maybe something else.'' The important characteristic is that they serve as the results of deliberative activity, indicating that options were considered and a decision was made. 

Remembering the results of deliberation is of little use unless you are committed to that decision. This is the second characteristic taken from Bratman and expanded on by Cohen and Levesque \parencite*{Cohen1990a}. By committing oneself to a future course of action, a person stabilizes their activity. As a consequence, they are better able to make predictions about their future states. More importantly, when a person tells others about their plans, those people can make predictions under the supposition that in most cases people see their intentions through. In this manner, predictability provides information about reliability and enables others to build trust in someone's performance. So, there is social pressure for a person to keep true to their publicly announced intentions. In addition, one person can use another person's plans to inform their own, which can enable coordination and cooperation across a seemingly infinite diversity of jointly pursued activities. 

Because intentions encode commitments to act according to earlier decisions, they facilitate a third characteristic, which is the ability to coordinate with one's future self. In practice, this means that intentions provide anchor points for reasoning about the future. If you decide to walk to the bakery, some plans (e.g., stopping at a neighboring store on the way) are made more feasible, whereas other plans are ruled out or constrained (e.g., your purchases are limited to what you can comfortably carry). Moreover, unrelated plans also take these commitments into account. For instance, intending to buy bread means that even before going to the bakery, you can plan to have a sandwich for lunch the next day. On a larger scale, when we consider long-term goals for our lives, we make plans like attending college, pursuing a major, and applying at specific companies all in service of that goal and all in relationship to each other even though those plans may not come to fruition for years. So, whether at the small scale or the large scale, intentions provide a means to organize our lives over time. A primary value of intending is that it biases people toward preserving the results of their earlier deliberation.

\subsection{Intention Through the Lens of Action}
Whereas Bratman's philosophical analysis emphasizes the role of intention in planning, Malle and Knobe \parencite*{Knobe1997} set out to determine how intentions and intentional actions are related in our folk-psychological intuitions. As a starting point, Bratman, Malle, and Knobe would all agree that a causal history connecting an intention with a specific action does not necessitate that the action was intentional. For instance, suppose that someone playing their first game of darts intends to hit the bullseye on every throw of a round. The first dart hits, and the second and third fly wide of their target. Did the person intentionally hit the bullseye? Speaking strictly in terms of causal history, they did, but because we know that the person lacks the \textit{skill} required to hit their mark regularly, we may attribute the bullseye to beginner's luck or call it a fortunate accident. Confirming this intuition, Malle and Knobe's experiments found that people see skill, the ability to reliably follow through on an intention, as an important component of intentional action. 

Additionally, Malle and Knobe's experiments identified \textit{awareness} as an expected characteristic of intentional action. In their model, awareness refers to an agent knowing that it is fulfilling the intention through its action. To illustrate, consider the following two scenarios. Suppose that Amelia decides to give away her old winter coat, puts it in a box, and takes that box to a donation center. Clearly, she intentionally donated her coat. Compare that to a case where Amelia's mom finds the coat, puts it in the box, and heads to work while Amelia is collecting other items. After searching for the coat unsuccessfully, Amelia gives up and takes the box to the donation center. When she returns home, she asks her mom about the coat and discovers that she had already donated it. Although Amelia intended to donate the coat, she was not aware that she was doing so at the time. Note that the limited scope of this definition of awareness means that we do not have to fully determine what awareness is. Instead, we might only stipulate that an agent can report which intentions it is carrying out at a particular time.

The awareness condition of intentional action is easy to confuse with other phenomena, such as when our actions have unintentional consequences. We may not have known that an outcome was possible or we might have been careless in our execution. For example, someone might desire to read a book, intend to reach for it, and knock over a cup of coffee in the process. Reaching for the book was intentional, but spilling the coffee was not. How can the same action be both intentional and unintentional? Anscombe \parencite*{anscombe79}  accounted for these situations by saying that actions are intentional \textit{under a description}. For instance, the person's action may have been intentional under the description reaching-for-the-book but unintentional under the descriptions knocking-over-the-cup and extending-an-arm. The lack of awareness is not what makes knocking-over-the-cup unintentional, the action under that description was never intended in the first place. Of course, if every time a person reached for a book, they knocked over a cup, we might suspect that they were using the book as excuse for being a nuisance. Therefore, an agent should be clear and accurate in describing their intentions so that others can continue to trust them.

\subsection{Intention Through the Lens of Control}
Although folk psychology usually refers to the relationships among intentional and other cognitive states, people also have expectations about how these states are manifested. Pacherie \parencite*{Pacherie2009} and Mylopoulos and Pacherie \parencite*{Mylopoulos19} present one way of describing this link between cognition and action that centers on three categories of intention: distal, proximal, and motor. That model arranges these forms of intention in a cascade with distal intentions sitting at the top and playing a role in planning (cf. Bratman's future-directed intentions). The distal intentions are representationally abstract and influence the selection of proximal intentions, which are situated and concrete. This process involves grounding conceptual information from the distal intention with perceptual information. Proximal intentions, in turn, parameterize motor intentions, which guide the selection of an appropriate motor program. As an example, you might form a distal intention to grab a pen from your desk. The corresponding proximal intention would pick out a specific pen, and the motor intention would select a motor program for grabbing it. In addition to providing a connection between cognition and action, Pacherie's model also allows for an exchange of information across representational levels. For instance, a failure to grab a pen could lead to executing a different motor program, selecting a different pen, or replanning using a new performance constraint. 

Along these lines, if someone is driving and finds that their regular route is blocked, we expect them to find a new path. If a person is writing software and realizes that their programming skills are inadequate, we expect them to learn new material or seek assistance. Likewise, if a robot finds a door it cannot open, we would expect it to find another way of reaching its goal whether that involves acquiring a new skill, constructing a tool, seeking assistance from nearby people, or creating a new plan to move forward. These abilities imply that intentions contain processes for guiding and monitoring activities with each category of intention playing a distinctive role \parencite{Pacherie2009}. Moreover, we expect that people will avoid plans that are unachievable (i.e., with no possibility of guidance) and will either avoid or discount plans where they cannot effectively monitor progress. In short, intending to do something without also believing that you can control yourself or others in the necessary way through guidance and monitoring is irrational.

Guidance and monitoring processes work, in part, by drawing our attention to information relevant to our current intentions. Watzl identifies attention as the reshaping of priority structures that arrange occurrent, subject-level, mental states, and he writes that intentions ``constrain how [a subject's] priority structures are going to evolve in the future'' \parencite*[][p. 143]{watzl2017}. Similarly, Wu, who views attention as ``selection for action,'' writes that ``the subject is attuned during [intentional] action to relevant information such that it is deployed to inform the subject's response'' \parencite*[][p. 98]{Wu2016}. In other words, intentions configure attentional priorities which then control behavior. From a folk-psychological perspective, we seem to know this when we talk about distractions that pull us off task and failures that occur while multitasking. Consider someone driving down the highway who decides to adjust their radio to a new station. Suppose the driver misses their exit even though it was clearly marked and their GPS was providing directions. In this case, visually attending to the radio guided the pressing of buttons or turning of knobs, aurally attending to the music enabled monitoring for content and volume, and those attentional priorities led to missing the sign for the exit and the instructions from the GPS. The driver's intention to change the radio station overrode the attentional priorities that would have kept them on route. 

The driver in this example might have avoided missing the exit if they had used what Gollwitzer \parencite*{Gollwitzer1999} calls \textit{implementation intentions} as protection. Implementation intentions are small plans that look like simple rules. For instance, someone on a diet might say, ``When I order dinner at the restaurant, I will skip dessert.'' Even if that person forgets their intention throughout the day, opening the menu may trigger the memory of the earlier commitment. Similarly, the driver could have said, ``When I am near my exit, I will make sure to listen to the GPS.'' As with plans, these simpler intentions are commitments we make to ourselves, but instead of giving step-by-step guidance, they tune our behavior as we engage in some broader activity. Thinking back to folk psychology, this ability means that we expect that intentions can have triggering conditions and can be combined in ways that tune our actions in the moment.

\section{Progress Toward a Computational Folk-Psychology}
Recall that the purpose of examining intention is twofold. First, gaining familiarity with the depths to which others have analyzed intentional states reveals the limitations in existing representational approaches. Second, elaborating on the characteristics of intention found in folk psychology and summarized in Table~\ref{tab:fpi} provides a starting point for a discussion about evaluation and progress. To the first point, consider how the elements in Table~\ref{tab:fpi} map onto a program like AlphaGo. Viewed through the lens of action, the program appears to have some aspects of intention in that it clearly has the necessary skills, it can report what action it is taking (here, the action and the report are identical), and it is clearly only acting under the description of making moves appropriate to the game of Go. However, through the lenses of planning and control, there is no obvious mapping. AlphaGo commits to its current action not a future one, does not store deliberations that affect future play, and does not coordinate with its future actions. There appear to be no representations of intentions that separately enables reasoning and feedback across abstract, situationally grounded, and motor-program levels. Each move is made in isolation, so there is no time-extended strategy that requires guidance and monitoring, and there are no triggering conditions that result in pre-planned or stereotypical behaviors. This is a brief and informal analysis of one system, but to the second point, how in general should we determine whether a system aligns with our folk-psychological intuitions and how do we define progress?

\begin{table}[t]
    \centering
    \caption{Functions and properties of intentions from a folk-psychological perspective}
    \vspace{1ex}

    \label{tab:fpi}
        \begin{tabular}{|@{\extracolsep{\stretch{1}}}|p{4cm}|p{8cm}|@{}|}
        \hline
        \multirow{3}{2cm}{Planning} & \sr Store the results of deliberation         \\ \cline{2-2} 
                                  & \sr Commit to future action                   \\ \cline{2-2} 
                                  & \sr Coordinate with future self               \\ \hline \hline
                                  
        \multirow{3}{*}{Action}   & \sr Assume skill to carry out intention       \\ \cline{2-2} 
                                  & \sr Assume current intention can be reported  \\ \cline{2-2} 
                                  & \sr Require clarity and truth in report       \\ \hline \hline
        \multirow{3}{*}{Control}  & \sr Incorporate multi-level representations   \\ \cline{2-2} 
                                  & \sr Include guidance and monitoring processes \\ \cline{2-2} 
                                  & \sr Include triggering conditions             \\ \hline
        \end{tabular}
        
\end{table}

Taking the extensive work on autonomous driving as an example, we can begin to think about how evaluation of intentional states might be standardized. In a report by the National Highway Traffic Safety Administration \parencite[NHTSA;][]{ndot2018}, there is a list of categories of behavioral competencies for driving that are subdivided into specific ones. For instance, tactical maneuvers like parking are broken down into specific behaviors including navigating a parking lot, locating open spaces, and making appropriate parking maneuvers. Similarly, for \textit{object and event detection and response} capabilities the category of \textit{traffic control devices and infrastructure} includes following local driving laws and reacting to speed limit changes. These competencies are broad in scope because they are meant to guide the development of test scenarios. (Examples of these scenarios are too detailed to include here but are available in Appendix~B of NHTSA's report.) In the case of intentions, Table~\ref{tab:fpi} and future extensions to it are meant to serve the same purpose. One evaluation scenario could combine committing to future actions with assumptions that the tested system has the skill to execute those actions and can monitor their performance for success. The test scenario will need to state how these capabilities connect with the exact tasks or behaviors that the system is meant to carry out. 

On a pessimistic side, autonomous driving research has also shown that the more we know about how to implement driving, the more complicated and detailed the activity appears to be. There are seemingly endless corner cases, whether they involve challenges to computer vision such as treating the moon or restaurant signs as traffic lights \parencite{levin21} or control failures such as attempting to turn into oncoming vehicles or light-rail lines \parencite{hammer22}. Many of the problems that arise are not about driving per se but are instead tied to the broader set of challenges related to acting in an open-ended environment that is suffused with dynamicity. Driving in the world, it turns out, is not like playing shogi, go, or chess. In a different sense, implementing intentional states is not like building an autonomous car. Parking is not believing. For instance, we know how to determine if a car is properly parked within a space, but how do we determine whether a belief is properly held by an agent? If the functional roles of intentional states differ situationally, then the challenges are even greater. We know little about what ``implementing intention'' might require compared to addressing task-centered domains like driving. That is not to say that intention, belief, desire, and hope are as nebulous as intelligence. But, to the extent that intentional states are moving targets forever on the horizon, it may be necessary to rethink what a progressive research program looks like. 

My suggestion would be to adopt the apophatic methodology originally developed for a science of consciousness \parencite{Bridewell2021}. As a topic of scientific research, consciousness is resistant to objective measurement. Consequently, scientists develop proxy measures to target in their evaluations. For instance, one might say that a robot is conscious because it passes a mirror test \parencite{torigoe2009} similarly to how one might pronounce that a program is intelligent because it wins chess games \parencite{Feigenbaum1963}. These kinds of claims beg the question of consciousness or intelligence (i.e., intelligence just is the ability to play a board game), and people are right to respond with incredulity and doubt. Apophatic science offers an alternative route for progress. 

The core of this methodology should be familiar as it draws from Newell's \parencite*{newell73} prescriptions for progress in cognitive science. The central idea applied to AI is to implement a series of computational models such that each one accounts for more intelligence-relevant phenomena than its predecessor. Additionally, the fidelity to mechanisms that enable human intelligence, according to the researchers developing the models, should also increase. Comparatively, the validation step of the apophatic methodology is radical. Once a model meets its original design criteria, it is recognized as inadequate---the criteria are insufficient for capturing intelligence computationally. The next step is to ask, ``What's missing?'' Given an answer, the set of criteria are extended, and the expanding list of properties and functions collects the not-quite-it of intelligence as the potentially unending process, summarized in Table~\ref{tab:asp}, continues. 

\begin{table}
\caption{The apophatic methodology in a nutshell.}
\label{tab:asp}
\begin{framed}
\noindent
Given a phenomenon $P$ and an initial set of functional roles $X$
\\ \textbf{loop} $(X)$
\\ \hspace*{1em} $P = X$
\\ \hspace*{1em} Implement $P$ in model $M$
\\ \hspace*{1em} Verify that $M$ is a correct implementation of $P$
\\ \hspace*{1em} Reject $M$ as insufficient
\\ \hspace*{1em} Identify functional role $a$
\\ \textbf{recur} $(X \cup a)$
\end{framed}
\end{table}


Even if intentional states turn out to be easier to objectively define than intelligence or consciousness, the discussion in Section~\ref{sec:lookia} suggests that they are not circumscribed as what we find in current implementations (e.g., as plans or policies). The good news is that, from a practical standpoint, only a subset of the known functional roles and properties may be required by any individual cognitive system. Nevertheless, by methodically exploring intentions, beliefs, and other intentional states we increase our ability to identify which properties and functional roles will satisfy the needs of any distinct task. To illustrate this point, the success of AlphaGo against human challengers suggests that including other functional roles of intention is unnecessary to meet the system's purpose. If the program were a multifunctional game player \parencite{Langley2017} that was required to justify its moves by appeal to deliberation (``Why did you place that piece?''), then more work would be required.

For the apophatic method to operate effectively, it is important to embrace stakeholder involvement. By this, I mean not only that we should collaborate with researchers in other disciplines who have important contributions to make regarding intentionality but also that we should engage with the communities that would benefit from the systems we build. This level of interaction has taken place in the realm of autonomous vehicles, with people providing information on successful drives and identifying challenging corner cases. We have also seen this advice being given by groups working on DARPA's XAI program \parencite{ihmc2021}. People approach software with different needs and goals in mind and a one-size-fits-all explanation is not going to be sufficient in all scenarios. As Hoffman and colleagues  note, ``By hearing first-hand from the different stakeholders about what they need in terms of explanations, developers will be better able to help stakeholders develop good mental models of a system'' \parencite*[][p. 3]{ihmc2021}. When it comes to intentions, a planning system that is not connected with execution may not need to commit to a future action, but a user would certainly expect it to store the results of its deliberation. Compare that case with a joint planning and execution system where users would expect it not only to store its plan but also to treat those results as commitments to the planned actions. 





As may be obvious by now, the questions ``What is intention and how do we model it?'' are too large to answer in this paper. That task falls to the research community in collaboration with other stakeholders. Undoubtedly multiple working groups will be needed, each focusing on a facet of folk psychology (e.g., intention, knowledge, belief, and desire) and meeting to ensure agreement and coherency across high-level concepts. Overall, the goal should be to reduce ambiguity when describing intentionality. By this, I mean that whenever researchers claim that their systems have beliefs or intentions, they should be able to point to specific properties and functional roles, tie those to the tasks that they are addressing with their systems, and explain why these are sufficient.  

\begin{figure}
    \centering
    \includegraphics[width=5in]{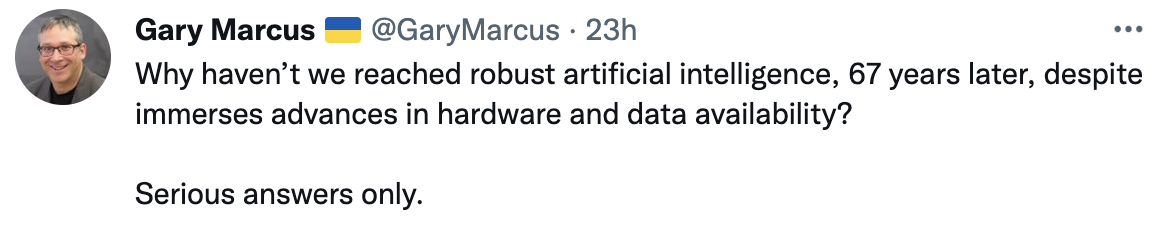}
    \caption{A recent post on Twitter by Gary Marcus \parencite*{marcus22}, a New York University Professor Emeritus and founder of two AI companies. The responses he received were varied and although anyone was free to reply, the contents reflected those in the AI community. (Note: 'immerses' is a typo for 'immense'.)}
    \label{fig:marcus}
\end{figure}

There is often considerable handwringing about how progress in AI is measured and why the field fails to meet some long-established goals. Figure~\ref{fig:marcus} provides only one of the more recent and public examples of this conversation. To address that question for an entire discipline covering almost 70 years of effort is too much to ask. As an alternative, I have attempted to provide a response for one aspect of AI that is especially relevant to cognitive systems. My hope is that instead of contributing to some debate regarding who or what is at fault, I have illustrated the scale of the challenges we face, pointed to recent advances in evaluative approaches, and indicated a method for measuring scientific progress in AI. 

\begin{acknowledgements} 
    \noindent
    This paper emerged from conversations with Paul Bello and has benefitted from the suggestions of Andrew Lovett. The author's effort was supported by the Office of Naval Research through the Science of Artificial Intelligence program award number N0001422WX00033. Distribution Statement A. Approved for public release; distribution is unlimited.
\end{acknowledgements}

\emergencystretch=1em
\printbibliography
\end{document}